\documentclass[sigconf,nonacm,balance=false]{acmart}

\AtBeginDocument{%
	}

\setcopyright{none}
\usepackage{booktabs}
\usepackage{multirow}
\usepackage{graphicx}
\usepackage{enumitem}
\usepackage{url}
\usepackage{natbib}
\usepackage[table]{xcolor}
\usepackage{colortbl}

\setcitestyle{nocompress}

\usepackage[most]{tcolorbox}

\definecolor{PromptFrameColor}{HTML}{B778F6}
\definecolor{PromptBackColor}{HTML}{EFE1FD}

\newtcolorbox{PromptBox}[1]{%
	breakable,
	colback=PromptBackColor,
	colframe=PromptFrameColor,
	colbacktitle=PromptFrameColor,
	coltitle=white,
	title={#1},
	fonttitle=\bfseries\small,
	boxrule=0.5pt,
	arc=1mm,
	left=1mm,
	right=1mm,
	top=0.5mm,
	bottom=0.5mm,
	boxsep=0.5mm
}

\definecolor{WAACGreen}{HTML}{E5F2E5}
\definecolor{ChangeRed}{HTML}{C94C4C}

\definecolor{BestGreen}{HTML}{E2F0E2}
\definecolor{SecondOrange}{HTML}{FCE8D5}

\newcommand{\bestcell}[1]{%
	\begingroup
	\setlength{\fboxsep}{0pt}%
	\colorbox{BestGreen}{\strut\textbf{#1}}%
	\endgroup
}

\newcommand{\secondcell}[1]{%
	\begingroup
	\setlength{\fboxsep}{0pt}%
	\colorbox{SecondOrange}{\strut\underline{#1}}%
	\endgroup
}

\title{Who Schedules Agent Memory? A Sidecar Approach to Long-Horizon Tasks}

\author{Qijia Zhuang}
\affiliation{%
	\institution{Nanjing University}
	\city{Nanjing}
	\country{China}
}

\author{Rui Liu}
\affiliation{%
	\institution{Northeastern University}
	\city{Qinhuangdao}
	\country{China}
}

\author{Wenqi Hu}
\affiliation{%
	\institution{Nanjing University}
	\city{Nanjing}
	\country{China}
}

\author{Yuxiang Ren}
\authornote{Corresponding author.}
\email{renyuxiang@nju.edu.cn}
\affiliation{%
	\institution{Nanjing University}
	\city{Nanjing}
	\country{China}
}

\hypersetup{
	keeppdfinfo=true,
	pdfauthor={Qijia Zhuang, Rui Liu, Wenqi Hu and Yuxiang Ren}
}

\makeatletter
\def\@mkauthors{%
	\gdef\@currentauthors{}%
	\gdef\@currentaffiliation{}%
	\global\setbox\mktitle@bx=\vbox{%
		\unvbox\mktitle@bx\par\medskip
		\centering
		{\@authorfont
			Qijia Zhuang\textsuperscript{1},
			Rui Liu\textsuperscript{2},
			Wenqi Hu\textsuperscript{1},
			Yuxiang Ren\textsuperscript{1,*}\par
		}%
		\smallskip
		{\@affiliationfont
			\textsuperscript{1}Nanjing University, Nanjing, China
			\quad
			\textsuperscript{2}Northeastern University, China\par
		}%
		\bigskip
	}%
}
\makeatother

\begin{document}
	
	\title{Rethinking Agent Trajectory Compression: A Workflow-Aware Perspective}
	
	\begin{abstract}
		As long-horizon agents increasingly rely on multi-step tool interactions to accomplish complex tasks, their execution trajectories continuously accumulate structured outputs and intermediate artifacts, leading to growing context overhead. Existing compression methods primarily rely on text importance or fixed truncation rules, often overlooking cross-step dependencies within workflows and potentially discarding critical information required for subsequent task execution. To address this challenge, we propose \textbf{WAAC}, a workflow-aware framework for agent trajectory compression. WAAC structurally models successful historical trajectories to extract reusable workflow compression experience and incorporates online backward dependency analysis to dynamically identify information retention requirements. By combining current dependency constraints with historical experience, WAAC selects appropriate compression strategies for different types of historical information. Experiments on PinchBench and Odysseys demonstrate that WAAC achieves an overall score of 87.6 and a perfect completion rate of 19.0\%, respectively, while reducing execution costs by 19.7\% and 29.1\% compared with the strongest baseline methods. These results demonstrate the effectiveness of workflow-aware trajectory compression in improving task performance and cost efficiency for long-horizon agents. Code and resources are available at \url{https://anonymous.4open.science/r/waac-6527/}.
	\end{abstract}
	
	
	
	
	
	\maketitle

	\section{Introduction}
	
	With the continuous development of tool-using agents driven by large language models~\cite{claude_code, openai_codex, openclaw}, agents are expanding from single-turn question answering to long-horizon tasks that require multi-turn reasoning, external tool calls, and continuous environmental interaction ~\cite{waac_ref01, waac_ref02, waac_ref03, waac_ref04}. During web agent execution, interaction trajectories continuously accumulate ~\cite{waac_ref05, waac_ref06} as tasks advance, causing historical decisions, tool calls, environmental feedback, and intermediate results to continuously enter the model context, thereby increasing window occupancy, reasoning overhead, and execution latency ~\cite{waac_ref08, waac_ref09, waac_ref10}. For agents that need to run for a long time or continuously process multiple sub-tasks, effectively managing the continuously growing interaction history has become a key problem affecting their efficiency and scalability ~\cite{waac_ref11, waac_ref12, waac_ref13}.
	
	Agent trajectories differ from ordinary documents because their contents participate in subsequent computation. A tool observation may provide a descriptive fact, an exact argument for another tool, or a reference to an intermediate artifact. Removing certain key fields can leave the history semantically understandable but insufficient for the next operation. Conversely, repeatedly exposing the complete contents of an already stored artifact can consume context without providing additional information needed by the current step ~\cite{waac_ref14, waac_ref15}. The compression problem is to determine both which historical fields remain relevant and which representation is sufficient for each field. This requires relating information to its downstream use, rather than assigning a single importance score to an entire message or uniformly shortening all previous observations
	
\begin{figure}[t]
		\centering
		\includegraphics[width=1.05\linewidth]{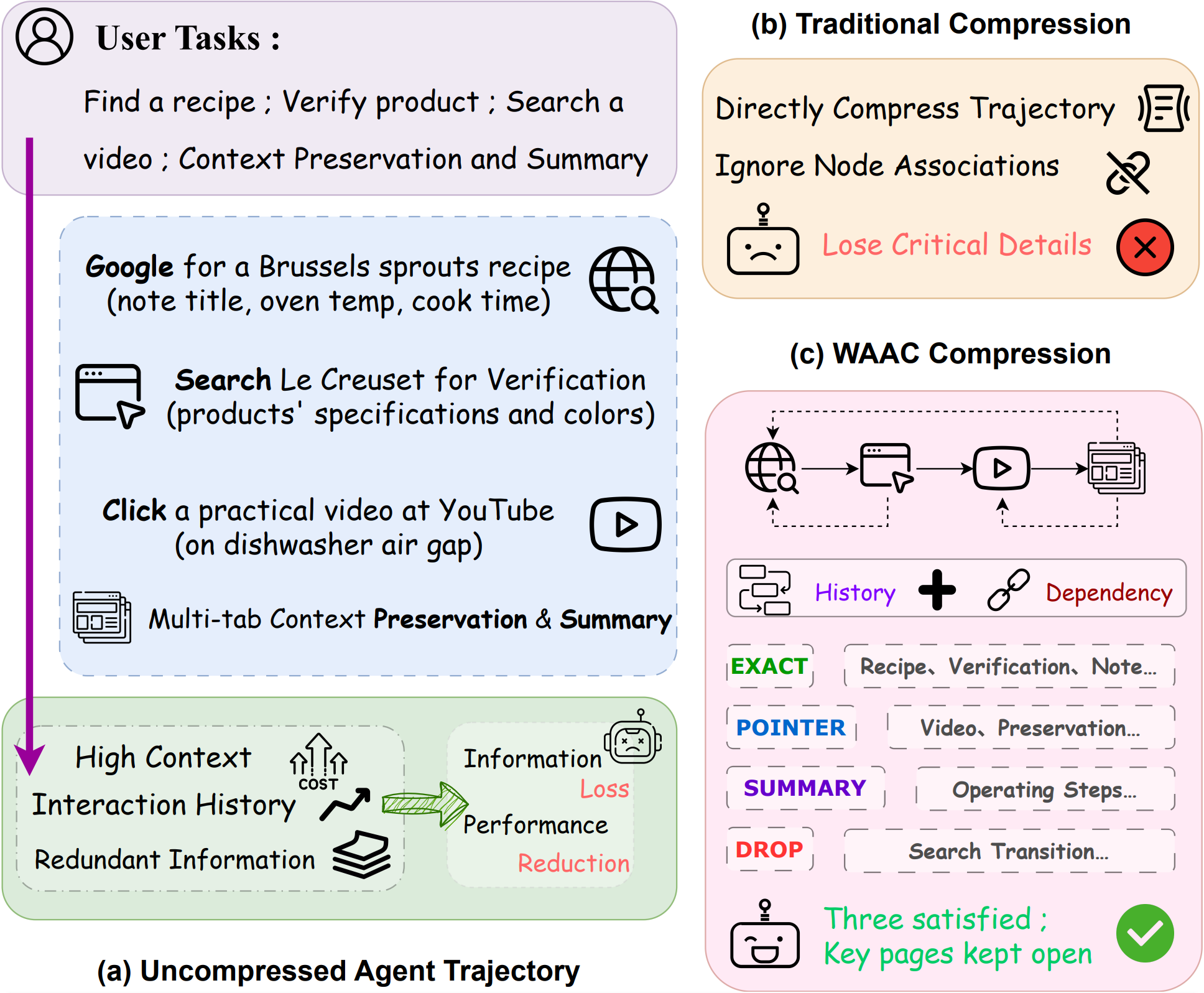}
		\caption{Motivating example. (a) Accumulating browsing history. (b) Loss of task-critical fields under compression. (c) WAAC assigns field-specific representations using workflow dependencies and historical policies.}
		\Description{(a) An uncompressed agent trajectory accumulates interaction history across multiple tool-assisted tasks, increasing context overhead and retaining redundant information. (b) Traditional compression may discard information required by subsequent operations when dependencies between trajectory nodes are not explicitly considered. (c) WAAC combines historical workflow knowledge with dependency-aware compression to apply different preservation policies, including EXACT, POINTER, SUMMARY, and DROP, to individual fields. This enables the agent to reduce redundant context while retaining information needed for subsequent execution.}
		\label{fig:intro}
	\end{figure}

	Existing methods alleviate resource overhead from perspectives such as text redundancy, reasoning state, historical storage, or context length ~\cite{waac_ref17, waac_ref18, waac_ref19, waac_ref20, waac_ref21}, but when the compression object becomes continuously growing agent trajectories that contain tool calls and structured intermediate artifacts, they still face a key challenge (Figure \ref{fig:intro}(b): how to determine which information in the history is still depended on by subsequent operations, and at what fidelity this information should be retained? Relying solely on text importance, fixed truncation rules, or independent memory retrieval may not accurately characterize the dependency relations between fields and future operations; meanwhile, historically successful trajectories may already contain reusable workflow structures and effective compression experience ~\cite{waac_ref16}. How to jointly utilize current execution dependencies and historical workflow priors constitutes a problem that long-horizon agent trajectory compression needs to further solve.
	
	Based on the above observations, this paper proposes an innovative agent compression architecture. Its core idea is that \textbf{effective trajectory compression needs to explicitly analyze the dependency of the current workflow on historical fields and utilize the reusable experience in successful historical trajectories to guide compression decisions}. In the offline stage, successful trajectories are converted into Typed DAGs, which structurally represent operations and field dependencies, and construct matchable historical workflow fragments and their field-level compression policies; in the online stage, backward dependency analysis is used to determine the information retention requirements needed for current execution, historical fragments are matched to obtain reusable policies, and the final compression decision is formed through conservative fusion. While compressing the model-visible context, the runtime execution state remains unchanged. By combining structured representation of workflows, current dependency constraints, and reuse of historical experience (Figure \ref{fig:intro}(c)), it aims to reduce unnecessary historical information overhead while lowering the risk of decision-critical information being over-compressed.
	
	The main contributions of this paper are as follows: 
	\begin{enumerate}[label=\textbullet, leftmargin=*, itemsep=6pt, topsep=4pt]
		\item Proposing a \textbf{workflow-aware agent trajectory representation and field-level adaptive compression framework}. Trajectories are represented as Typed DAGs to explicitly characterize field-level data dependencies among operations, and adaptive fidelity control is applied to fields based on dependency constraints and historical experience (covering different retention strengths such as \texttt{DROP}, \texttt{SUMMARY}, \texttt{POINTER}, and \texttt{EXACT}). 
		\item Designing a compression mechanism that combines \textbf{online dependency analysis with offline historical matching}. Backward dependency analysis of the current trajectory is combined with matching of successful historical workflow fragments to coordinate the information retention requirements needed for current execution with historically reusable compression experience.  
		\item Proposing a design perspective that measures the retention value of historical information by its subsequent uses in the workflow. Trajectory compression is characterized as \textbf{a field-level fidelity allocation problem oriented to execution requirements}, establishing a clear connection between information retention and operational dependencies, and providing a structured basis for context management of long-horizon agents.
	\end{enumerate}
	
	\section{Related Work}
	
	\paragraph{Agent-oriented trajectory compression.}
	Long-horizon, sustainable agents are becoming prevalent in real tasks such as Web Search~\cite{wei2025browsecomp, jang2026odysseys}, and the continuous growth of interaction history makes trajectory-level compression a critical problem. Early works mostly start from general LLM efficiency, performing token/context-level deletion and rewriting on relatively static prompts or documents~\cite{pan2024llmlingua, li2023compressing}. On the system side, KV cache compression has also developed in parallel to reduce inference memory and latency~\cite{li2024snapkv, cai2024pyramidkv}; its trade-offs are usually driven by attention or caching strategies and do not prescribe which trajectory semantics should be retained from the perspective of agent decision dependencies. Meanwhile, memory mechanisms alleviate window pressure through external storage, with their focus on read/write and retrieval~\cite{zhong2024memorybank, chhikara2025mem0}. When the object becomes continuously growing agent trajectories that contain tool dependencies, relying solely on any of the above routes is difficult to directly answer “which parts of the history can be safely compressed”~\cite{ye2025agentfold, feng2026agentocr}. Unlike compression that mainly shortens a single input, agent trajectory-level compression focuses on: in multi-turn tool interactions, which critical information from preceding steps may still determine subsequent decisions, thereby making trade-offs between length reduction and continued executability.
	
	\paragraph{Experience reuse of agent trajectories.}
	The success of multi-step tool agents often depends on repeatable sub-processes~\cite{ding2026atomic}, such as similar retrieval paths and tool call sequences, and thus historical trajectories accumulate transferable work experience. One line of work writes interactions into external memory and retrieves relevant segments in new tasks to assist decision-making~\cite{packer2023memgpt}; another line induces reusable skills, workflow templates, or experience rules from successful or failed trajectories, shifting reuse from indexing raw records to invoking more abstract process knowledge~\cite{zhao2024expel, wang2024agent}. On this basis, experience reuse oriented to agent trajectories themselves further aligns transferable segments in historical trajectories directly through structural modeling and matching, thereby more fully utilizing the process information contained in the trajectories~\cite{zheng2026evaluating}.
	
	\section{Methodology}
	
	\subsection{Overview}
	
\begin{figure*}[t]
		\centering
		\includegraphics[width=\textwidth]{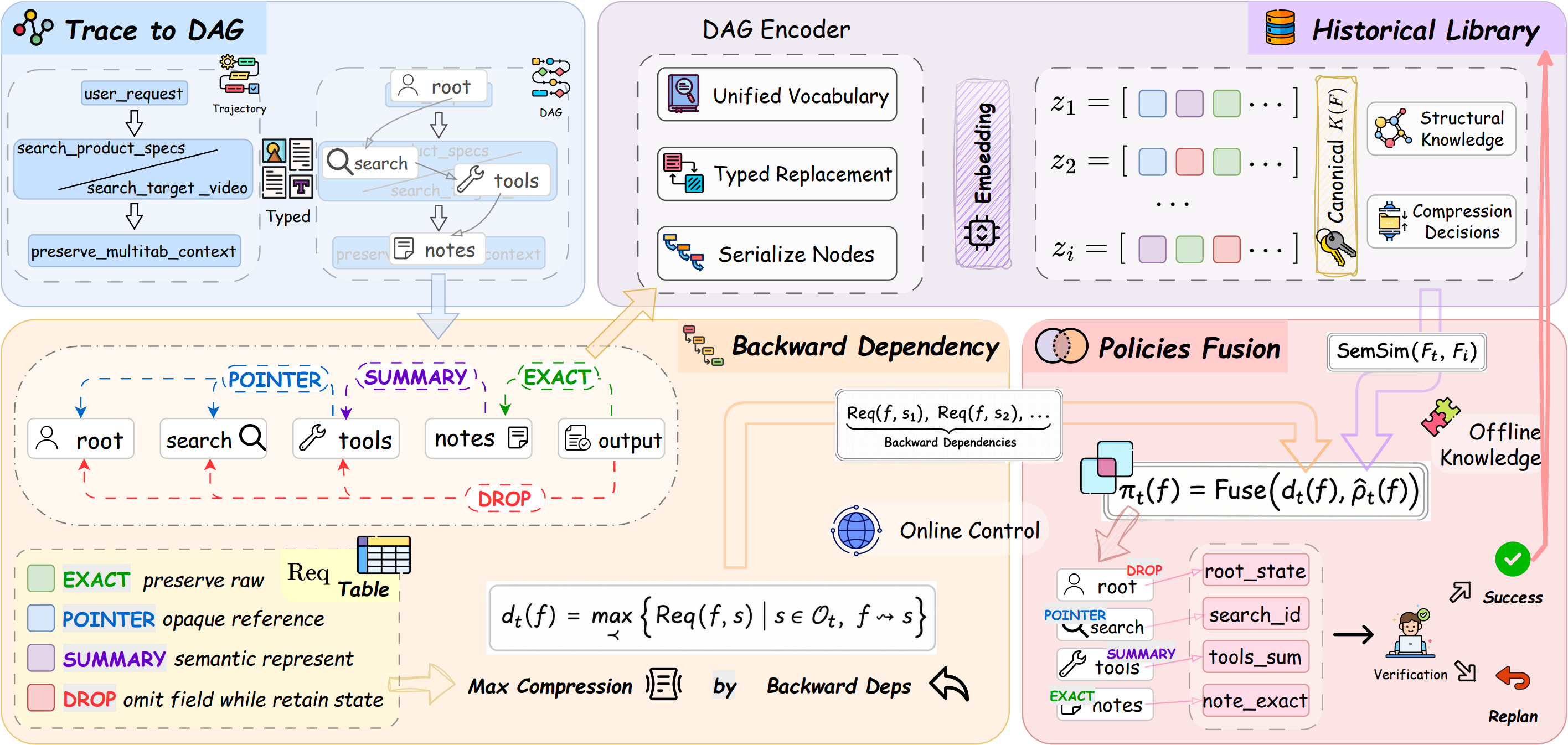}
		\caption{Overview of WAAC. Offline processing converts verified development trajectories into canonical workflow fragments and field-level policy records. Before each online model invocation, WAAC derives requirements from outstanding obligations, retrieves compatible historical policies, and combines them through a conservative join. The resulting projection changes the model-visible history while retaining original events and artifacts in runtime storage.}
		\Description{The Framewrok provides a detailed explanation of the WAAC architecture’s processing of converting traces into DAGs, the management of the History trajectory library, the analysis and computation of online dependencies, and the final strategy fusion selection.}
		\label{fig:framework}
	\end{figure*}

	As shown in Figure \ref{fig:framework}, WAAC constructs a field-level projection of the interaction history before each model invocation. At step \(t\), it takes the observable trajectory prefix \(H_{\leq t}\), retained runtime state \(S_t\), outstanding execution obligations \(\mathcal O_t\), and a verified history library as input. Historical matching first initializes field policies and supplies
	reusable dependency-analysis results. Online analysis then validates
	their applicability, recomputes changed or uncovered requirements,
	and conservatively combines the current requirements with the
	initial proposals. Details are provided in
	Appendix~\ref{ap_history_reuse}. The resulting policies determine which fields are exposed exactly, summarized, represented by recoverable references, or omitted from the model-visible context. Original events and runtime artifacts remain available in \(S_t\). WAAC aims to reduce visible-history tokens while satisfying the requirements represented by its registered schemas and dependency graph.
	
	\subsection{Construction of Typed Trace-to-DAG and Offline History}
	\label{sec:offline}
	
	\subsubsection{Typed trajectory representation.}
	
	To ensure the stability of the Agent execution path and subsequent trajectory optimization, we introduce the construction of a DAG to represent the Agent Workflow. The user instruction serves as the root node of the DAG, and the nodes and edges of
	$G = (V, E)$ are defined as:
	
	\begin{equation}
		V = \{ v_i = (op_i, I_i, O_i) \}, \quad E = \{ e_{ij}^{p \to q}: O_i[p] \to I_j[q] \}
	\end{equation}
	
	Each node \(v_i=(op_i,I_i,O_i)\) represents an observed operation together with its input and output fields. The indices \(i\) and \(j\) identify operation nodes, whereas \(p\) and \(q\) denote fields within their outputs and inputs, respectively. Accordingly, an edge \(e_{ij}^{p\to q}\) records that the value produced at \(O_i[p]\) is consumed at \(I_j[q]\). This field-level representation distinguishes information required by subsequent operations from descriptive context that need not be preserved verbatim. For example, a tool result may contain an artifact path, a status value, and a lengthy report body. A subsequent operation may require the path and status explicitly while accessing the report body through the stored artifact. \textbf{Modeling these fields separately therefore allows their preservation policies to differ}.
	
	The graph is constructed from executed events rather than a predefined cyclic workflow template. Structured tool arguments and results are parsed using registered schemas, and each extracted field retains a reference to its source event and original value. A data edge is established when an explicit argument binding or a schema-supported reference links an earlier output field to a later input field. For example, a path returned by a file-writing operation can be linked to the corresponding path argument of a subsequent validation call. For unstructured records, an auxiliary extractor proposes candidate fields and relations, which are checked against the available evidence. Relations that cannot be reliably resolved are marked as uncertain rather than treated as confirmed dependencies; affected fields are conservatively retained \texttt{EXACT} to avoid unsafe compression.
	
	Beyond data dependencies, the graph captures execution semantics and control constraints. Data edges represent producer–consumer bindings between fields, whereas control edges represent observed conditions governing whether an operation is executed. Backward traversal can therefore recover both the values consumed by an operation and the conditions required for its execution. Execution order is tracked separately from data dependency: an earlier observation need not remain visible merely because it precedes the current operation. Retries and artifact updates are represented as distinct events, with dependency edges respecting event order to maintain an acyclic graph. Each node retains its operation and tool types, inputs, outputs, and execution or verification status, while artifact references distinguish object versions. This event-level representation provides a common structural basis for backward dependency analysis and historical fragment alignment.
	
	\subsubsection{Normalization and typed merging.}
	
	Normalization converts successful trajectories into canonical fragments $F_\tau$ so that equivalent workflow structures can be compared across task instances. We apply three deterministic transformations: mapping operations to a shared vocabulary, replacing instance-specific values with registered type placeholders, and serializing the resulting nodes and edges in canonical topological order. The transformation \textbf{preserves field roles and dependency mappings}, since these determine how information is used by subsequent operations.
	
	For example, \path{C:\data\a.xlsx} and \path{/home/user/b.xlsx} can both be represented by the type `WorkbookPath`. Their concrete values remain in the runtime artifact store, while the canonical fragment records which operation produces or consumes the corresponding path field. This separation allows historical compression policies to be associated with recurring workflow patterns while the current execution continues to use its own values and artifacts. The normalized representation provides the comparison key; it does not replace the executable state of the current task.
	
	\textbf{Only trajectories that pass task-level verification are inserted into the history store.} A history entry contains the canonical fragment, verification metadata, and the field-level policies observed in that trajectory. It does not contain concrete executable actions or external side effects.
	
	For each field $f$, we adopt an ordered policy set (from left to right, policies are ordered by increasing priority):
	
	\begin{equation}
		\texttt{DROP} \prec \texttt{SUMMARY} / \texttt{POINTER} \prec \texttt{EXACT}.
	\end{equation}
	
	The specific meanings of the policies are:
	\begin{itemize}[leftmargin=*, itemsep=2pt, topsep=3pt]
		\item \textbf{$\texttt{EXACT}$}: retain the original field value;
		\item \textbf{$\texttt{POINTER}$}: provide an opaque reference to the runtime object;
		\item \textbf{$\texttt{SUMMARY}$}: share a bounded semantic representation;
		\item \textbf{$\texttt{DROP}$}: omit the field from the visible context while retaining it in the runtime state.
	\end{itemize}
	
	Development trajectories are collected from the development split using initial compression policy and library initialization. At each compression checkpoint \(u\) in trajectory \(\tau\), we record the observable fragment \(F_{\tau,u}\), its canonical field correspondence, and the policy $\rho_t(f)$ actually applied after representation validation and fallback. The fragment and policy record therefore describe the state available at that checkpoint.
	After task-level verification confirms success, these records are admitted to the history library and aligned to canonical fields. 
	
	For records sharing the same canonical key, policies are aggregated by a conservative join:
	
	\begin{equation}
		\rho_t(f)=
		\max_{\prec}
		\left\{
		\rho_{\tau,u}(f)
		\;\middle|\;
		\tau\in\mathcal T_{\mathrm{verified}},
		\ K(F_{\tau,u})=K
		\right\}.
	\end{equation}
	
	The canonical key $K(F_\tau)$ utilizes deterministic hashing to normalize nodes and edges for identifying the workflow structure. The maximum selects the highest-fidelity policy observed for that field. When compressing this node, it is necessary to ensure that the operation has only a minor impact on subsequent nodes; adopting the maximum-fidelity policy avoids necessary information of one node being affected by the compression operations of other nodes, thereby reducing subsequent multiple data accesses and the substantial extra cost of recovering original data.
	In the case of Agent retry and reuse, other redundant branches are subjected to the $\texttt{DROP}$ operation according to the final successful path.
	The final storage is a typed policy cache:
	
	\begin{equation}
		K(F) \longmapsto \bigl( F, \rho_t, \text{verification} \bigr).
	\end{equation}
	
	It caches structural knowledge and compression decisions, corresponding to the successfully processed DAG sequences.
	
	\subsection{Online Dependencies, Matching and Compression}
	\label{sec:online}
	
	\subsubsection{Maximum compression determined by backward dependencies.}
	
	At step \(t\), let \(\mathcal O_t\) denote the outstanding execution obligations identifiable from the observable trajectory prefix. These obligations include pending tool inputs, explicit planned operations, final answer requirements, and verification checks. For backward analysis, each obligation is linked to the observed fields it references. These links represent current demands and introduce no future observations. For a field $f$, let $f \rightsquigarrow s$ denote that $f$ can reach the backward field $s$ via a data or control edge path in the current DAG. The function $\operatorname{Req}(f, s)$ specifies the minimum fidelity required by the backward field $s$.
	
	We compute the current dependency requirement of a field through reverse traversal from the backward segment:
	
	\begin{equation}
		d_t(f) = \max_{\prec} \bigl\{ \operatorname{Req}(f, s) \mid s \in \mathcal{O}_t,\; f \rightsquigarrow s \bigr\}.
	\end{equation}
	
	The reverse traversal determines which historical fields contribute to the current node tasks, while $\operatorname{Req}(f,s)$ determines the representation needed for each contribution. Reachability alone is insufficient: two obligations may depend on the same field but require different information from it. These requirements are specified by registered schemas and operation rules (Specific details can be found in Appendix \ref{ap_fusion}). If several obligations depend on the same field, their requirements are combined so that satisfying one obligation does not invalidate another. Fields with unresolved requirements are retained conservatively.
	
	The obligation set is derived from the observable execution prefix, including pending tool inputs, explicit plans, answer requirements, and verification checks. It does not assume access to future tool results. Because a new observation can complete an obligation, introduce a new one, or change the information required for verification, the \textbf{dependency requirements are recomputed as execution proceeds}. The same historical field requires exact retention at one step and become eligible for a reduced representation after its dependent operations have completed.
	
	\subsubsection{Historical fragment matching.}
	
	Historical matching identifies previously verified fragments whose field policies may be relevant to the current local fragment $F_t$. The system first queries the exact-key index using the canonical representation. \textbf{A candidate is eligible for reuse only when its typed operation structure, field roles, and action categories are compatible with the current fragment.} These checks distinguish reusable workflow structure from incidental similarity in task descriptions. Compatibility therefore establishes whether policies can be aligned to corresponding fields before semantic similarity is used to rank candidates (Specific details can be found in Appendix \ref{ap_offline}).
	
	For the current fragment $F_t$ and the set of historical fragments $F_i$ ($i \in \tau$), the comprehensive similarity is:
	
	\begin{equation}
		\text{FinalSim}(F_t,F_\tau) = \texttt{Max}(\text{SemSim}(F_t,F_i)).
	\end{equation}
	
	$\operatorname{SemSim}$ is the cosine similarity between normalized semantic descriptions of eligible fragments, and the maximum selects the highest-scoring compatible match. The semantic description captures workflow meaning for retrieval, while the canonical structure determines whether the proposed field correspondence is admissible. The selected entry contributes its field-level policy proposal $\widehat{\rho}_t(f)$; concrete artifact values and executable actions are supplied by the current trajectory. Historical policies are transferred only through validated field correspondences. Fields without an admissible correspondence receive no historical retention constraint. If no compatible candidate passes the acceptance threshold, the decision uses online requirements alone.
	
	\subsubsection{Conservative strategy fusion.}
	
	Let $\widehat{\rho}_t(f)$ denote the policy retrieved from an exact match or the highest-scoring compatible coarse match. The final policy is obtained by combining the historical proposal with the current dependency requirement:
	
	\begin{equation}
		\pi_t(f) = \operatorname{Fuse}\bigl( d_t(f), \widehat{\rho}_t(f) \bigr).
	\end{equation}
	
	Historical policies act as additional retention constraints. For a fixed observable prefix, runtime state, dependency graph, and obligation set, conservative fusion satisfies \(\pi_t(f)\succeq d_t(f)\). Historical retrieval therefore cannot authorize a lower-fidelity policy than the online requirement. Its potential benefit is to preserve context whose utility for planning or subsequent decisions.
	
	Fusion selects a representation that satisfies both the current dependency requirement and the retrieved historical policy. Under the preservation order, this is expressed as $\operatorname{Fuse}(a,b)=\max_{\prec}(a,b)$. SUMMARY and POINTER provide different capabilities: the former exposes bounded semantic information, while the latter supports access to an external artifact. When both capabilities are required and neither representation alone is sufficient, the conservative choice is EXACT. If the history is empty or no compatible fragment is found, the final policy reduces to $d_t(f)$. Historical experience therefore supplies an additional retention constraint; it does not relax a requirement established by the current workflow. Specific details can be found in Appendix \ref{ap_method}.
	
	The online requirements cover the demands identifiable from the observable prefix under the registered schemas; they are not a complete specification of all future uses. Historical records provide successful retention precedents for compatible workflows, while unresolved dependency evidence continues to trigger conservative retention. If no compatible history is available, fusion reduces to the online requirement.
	
	\textbf{The selected policies are applied to the model-visible context while the original trajectory and runtime artifacts remain available.} An exact field is exposed unchanged, a summary retains the required semantic content, a pointer identifies an externally recoverable object, and a dropped field is omitted from the visible representation. Pointer use requires a valid reference to the intended artifact, and uncertain recovery must fall back to exact retention. The same conservative principle applies when a reduced representation cannot be shown to satisfy the registered requirements. These conditions bound the preservation claim to the dependencies and schemas represented by the system.
	
	After receiving the compressed context, the agent chooses its next action and executes the corresponding tool. The resulting observation updates the graph and outstanding obligations for the next decision. A completed trajectory can contribute to the history store only after task-level verification confirms success, while unverified executions do not become successful policy precedents.
	
	Models propose trajectory structure or generate bounded textual representations. Requirement assignment, compatibility checks, and policy fusion are executed by deterministic code. Unresolved extraction outputs trigger conservative retention.
	
	\section{Experiment}
	
	In this section, we conduct a comprehensive empirical evaluation of WAAC on two representative Agent benchmarks. Specifically, our experiments aim to investigate the following three points: (\textbf{1}) the comparison of WAAC with other Agent architectures in terms of performance and token efficiency; (\textbf{2}) the performance of WAAC under long workflows and strong constraints; (\textbf{3}) the effectiveness analysis of WAAC’s components.
	
	\subsection{Experimental Setup}
	
	\paragraph{Benchmarks.}
	We evaluate WAAC on \textbf{PinchBench}~\cite{pinchbench} and\\
	\textbf{Odysseys} ~\cite{jang2026odysseys}. Both benchmarks induce continuously growing context. PinchBench covers multiple categories of real agent tasks including productivity, research, writing, programming, and analysis, where the agent must complete multi-step operations through tool calls and environmental feedback; under the continuous workflow setting, tasks are sequentially connected within the same session, and context accumulates with session length. In contrast, Odysseys focuses on long-horizon multi-site workflows on the real open web: the agent must continuously plan, navigate, and reason across multiple websites on the live Internet to accomplish complex goals such as cross-platform price comparison and itinerary planning, while handling large amounts of web observations and long-trajectory interactions, thus requiring the maintenance of more persistent cross-site context and more efficient execution strategies.
	
	\paragraph{Baselines.}
	
	We compare WAAC with compression methods (Keep-Last-N, LLMLingua-2~\cite{pan2024llmlingua}, TokenSkip~\cite{xia2025tokenskip}) and memory or trajectory-level context management methods (LCM~\cite{waac_ref09}, AgentFold~\cite{ye2025agentfold}, Mem0~\cite{chhikara2025mem0}, Summary). Among them, LLMLingua-2 and TokenSkip reduce context overhead from the perspectives of prompt compression and generation-side token skipping, respectively; Keep-Last-N only retains the most recent $N$ rounds of interactions as a truncation baseline; Summary performs periodic summarization of historical trajectories; LCM, AgentFold, and Mem0 correspond to lossless/structured context management, active folding-style trajectory compression, and external memory read/write, respectively.
	
	\paragraph{Implementation Details.}
	
	The main WAAC agent uses gpt-5.4-mini through the OpenAI-compatible Chat Completions API, with temperature 0, top-p 1, and a maximum output of 8,192 tokens per call. Structure extraction, bounded summarization, and semantic retrieval use embedding model(text-embedding-3-small), respectively. Requirement assignment, compatibility checking, and policy fusion are implemented by deterministic code.
	The shared-backbone comparison includes methods using the same task-executing model. These methods share task instructions, tool schemas, workspace initialization, stopping conditions, and scorers. Methods requiring dedicated checkpoints or inference backends are evaluated using actual checkpoints and adaptations. In particular, TokenSkip uses skipping rate and generation-side implementation.
	We evaluate 147 PinchBench tasks and 200 Odysseys tasks. The history library is constructed from development data and isolation rule and remains read-only throughout evaluation. Context budgets and baseline-specific settings are provided in Appendix \ref{ap_exper}.
	
	\subsection{Main Results}
	
	On PinchBench, WAAC achieves the highest Overall score among the compared methods, reaching \textbf{87.6} against 84.3 for LLMLingua-2 and 72.6 for Vanilla  (Table \ref{tab:pinchbench_main_results}). The reduction in resource consumption is mainly reflected on the input side: the input Token consumption of WAAC is 7.174 million, a decrease of \textbf{24.6\%} compared with Vanilla, while the execution time is shortened by \textbf{16.24\%}. Moreover, WAAC leads in task categories strongly related to trajectories such as “Research”, “Writing”, “Coding”, “Analysis”, “Log Analysis”, and “Memory”. 
	
	On the Odysseys benchmark, the “Perfect” completion rate of WAAC is \textbf{19.0\%} (AgentFold is 15.0\%), and the Rubric Avg is improved from 42.0 to \textbf{48.0} (see Table \ref{tab:odysseys_main_results}). Its “Partial” proportion is also increased from 38.0\% to \textbf{46.0\%}, indicating that the advantages of the method are not limited to fully completed tasks. Compared with the baseline model (Vanilla), the average number of execution steps of WAAC is reduced from 75 steps to 58 steps, and the trajectory efficiency is improved from 0.42 to \textbf{0.68}. Moreover, the cost is reduced by 40.1\% relative to Vanilla and by 29.1\% relative to the SOTA model. WAAC both improves task completion effectiveness and reduces input consumption or shortens the number of execution steps.
	
	\begin{table*}[t]
		\centering
		\caption{Performance and token/time consumption of PinchBench.}
		\label{tab:pinchbench_main_results}
		
		\renewcommand{\arraystretch}{1.2}
		\setlength{\tabcolsep}{5pt}
		
		\resizebox{\textwidth}{!}{
			\begin{tabular}{l c ccccccccccc c c c}
				\toprule
				
				\multirow{2}{*}{Method} &
				\multirow{2}{*}{Overall} &
				\multicolumn{11}{c}{Performance by Category $\uparrow$} &
				\multirow{2}{*}{Input(M)} &
				\multirow{2}{*}{Output(M)} &
				\multirow{2}{*}{Time (s)$\downarrow$} \\
				
				\cmidrule(lr){3-13}
				
				& & Prod & Res & Write & Code & Anal & CSV & Log & Meet & Mem & Skill & Integ
				& & & \\
				
				\midrule
				
				Vanilla
				& 72.6 & 66.2 & 44.4 & 71.0 & 72.4 & 66.6 & 77.0 & 82.8 & 75.6 & 92.6 & 54.4
				& \bestcell{88.1}
				& 9.496 & 0.403 & 15,705 \\
				
				Keep-Last-N
				& 78.0 & 75.4 & \secondcell{70.0}
				& 79.4 & \secondcell{88.0}
				& 78.6 & 74.3 & 77.5 & 82.9 & 87.5 & 60.8
				& \secondcell{85.8}
				& 9.312 & 0.353 & \bestcell{12,950} \\
				
				LLMLingua-2
				& \secondcell{84.3}
				& 87.0 & 48.1 & 87.2 & 87.6 & 82.7
				& \bestcell{91.9}
				& \secondcell{90.4}
				& \bestcell{88.3}
				& 86.2 & 67.2 & 78.9
				& 9.037 & 0.385 & 15,036 \\
				
				TokenSkip
				& 79.2 & 88.1 & 62.2 & 87.1 & 82.3 & 81.2 & 76.4 & 81.5 & 82.4 & 85.4 & 68.6 & 74.2
				& 9.458 & \bestcell{0.278} & 15,659 \\
				
				Summary
				& 81.4 & 73.3 & 62.6
				& \secondcell{87.7}
				& 87.1 & \secondcell{83.0}
				& 81.4 & 90.1 & 81.6 & 87.5 & 67.4 & 68.4
				& 9.220 & 0.337 & 16,128 \\
				
				LCM
				& 78.8 & 88.4 & 63.2 & 85.5 & 83.3 & 80.9 & 80.1 & 78.3 & 76.9 & 84.9
				& \secondcell{78.2}
				& 79.8
				& 8.945 & 0.298 & 18,881 \\
				
				AgentFold
				& 80.8 & \bestcell{90.8}
				& 69.4 & 85.2 & 84.6 & 82.9 & 83.5 & 86.2 & 71.5 & 88.1
				& \bestcell{78.4}
				& 76.1
				& \secondcell{8.927}
				& 0.348 & 17,421 \\
				
				Mem0
				& 78.2 & 67.4 & 67.8 & 84.0 & 83.5 & 81.2 & 82.1 & 84.6 & 70.1
				& \secondcell{96.7}
				& 75.3 & 73.2
				& 9.468 & 0.344 & 19,471 \\
				
				\midrule
				
				\textbf{WAAC}
				& \bestcell{87.6}
				& \secondcell{89.0}
				& \bestcell{75.9}
				& \bestcell{89.9}
				& \bestcell{89.1}
				& \bestcell{83.6}
				& \secondcell{89.5}
				& \bestcell{94.2}
				& \secondcell{86.4}
				& \bestcell{97.5}
				& 77.6 & 83.8
				& \bestcell{7.174}
				& \secondcell{0.285}
				& \secondcell{13,155} \\
				
				\bottomrule
			\end{tabular}
		}
	\end{table*}
	
	\begin{table*}[t]
		\centering
		\caption{Performance of different methods on the Odysseys task.}
		\label{tab:odysseys_main_results}
		
		{\setlength{\tabcolsep}{4pt}
			\renewcommand{\arraystretch}{1.0}
			\resizebox{\textwidth}{!}{
				\begin{tabular}{l ccccc c c c c c}
					\toprule
					
					\multirow{2}{*}{Method} &
					\multirow{2}{*}{O-M2W Judge $\uparrow$} &
					\multicolumn{5}{c}{Rubrics (\%) $\uparrow$} &
					\multirow{2}{*}{Avg. Steps $\downarrow$} &
					\multirow{2}{*}{Traj. Eff. $\uparrow$} &
					\multirow{2}{*}{Input(M)} &
					\multirow{2}{*}{Output(M)} \\
					
					\cmidrule(lr){3-7}
					
					& & Avg. & Perfect & Hard & Partial & Avg. Hit & & & & \\
					
					\midrule
					
					Vanilla
					& 12.0 & 31.0 & 7.5 & 1.8 & 22.0 & 1.9
					& 75 & 0.42 & 75.25 & 4.501 \\
					
					Keep-Last-N
					& 9.0 & 25.8 & 5.5 & 1.0 & 16.5 & 1.6
					& 84 & 0.31 & 84.52 & 5.046 \\
					
					LLMLingua-2
					& 8.6 & 23.5 & 4.5 & 0.7 & 14.1 & 1.4
					& 91 & 0.27 & 91.27 & 5.462 \\
					
					TokenSkip
					& 11.5 & 31.0 & 7.2 & 1.6 & 21.0 & 1.9
					& 75 & 0.43 & 75.34 & 4.551 \\
					
					Summary
					& 10.3 & 26.4 & 5.0 & 0.9 & 17.4 & 1.6
					& 82 & 0.34 & 82.64 & 4.924 \\
					
					LCM
					& 14.5 & 35.5 & 9.0 & 2.6 & 27.0 & 2.2
					& 65 & \secondcell{0.55} & 65.19 & \secondcell{3.481} \\
					
					Mem0
					& 7.5 & 23.9 & 8.3 & 2.0 & 19.3 & 1.5
					& \bestcell{54} & 0.38 & 62.86 & 3.903 \\
					
					AgentFold
					& \secondcell{17.0}
					& \secondcell{42.1}
					& \secondcell{15.0}
					& \secondcell{5.5}
					& \secondcell{38.0}
					& \secondcell{2.6}
					& 62 & 0.54
					& \secondcell{58.47}
					& 3.723 \\
					
					\midrule
					
					\textbf{WAAC}
					& \bestcell{20.5}
					& \bestcell{48.0}
					& \bestcell{19.0}
					& \bestcell{8.2}
					& \bestcell{46.0}
					& \bestcell{2.9}
					& \secondcell{58}
					& \bestcell{0.68}
					& \bestcell{41.49}
					& \bestcell{3.248} \\
					
					\bottomrule
				\end{tabular}
		}}
	\end{table*}
	
	\subsection{Ablation Study}
	
	To analyze the role of each component of WAAC, we remove the offline module and the online module separately under the same settings as the main experiments, and the results are shown in Table \ref{tab:ablation}.
	
	In the absence of structured trajectory compression, context accumulates rapidly with tool calls: even with truncation or summary-style heuristics, peak input and total tokens remain difficult to effectively constrain, and the Agent often continues to carry redundant history in long-horizon interactions. Full WAAC mitigates this trend through the fusion of online dependency constraints and offline segment priors, achieving 87.6 Overall on PinchBench while reducing Input, and latency to 7.17M, 13,155s; on Odysseys, it achieves 19.0\% Perfect (Rubric Avg 48.0), with average steps reduced to 58 and Trajectory Efficiency improved to 0.68.
	
	After removing the offline module (w/o Offline), the system retains only online constraints based on backward dependencies. On PinchBench, the average success rate drops to 81.8, Input rebounds to 7.882M (+9.87\%), and latency increase to 13,773s respectively; on Odysseys, Perfect drops to 15.5\%, and average steps rise to 63. This result indicates that dependency analysis of the current trajectory alone can preserve most task success, but compression becomes more radical—without the segment prior $\hat{\rho_t}(f)$ from successful trajectories, it is harder to proactively adopt more suitable compression levels on similar substructures, thus significantly weakening the efficiency advantage.
	
	Removing the online module (w/o Online) and using only offline matching leads to more significant degradation. On PinchBench, Overall drops to 80.6, Input rises to 7.643M (relative to Full +6.54\%), and Time reach 14,634s respectively; on Odysseys, Perfect further drops to 12.7\%, and average steps remain around 61 but Trajectory Efficiency declines notably. This indicates that without the field-level safety boundary provided by $d_t(f)$, historical similarity may misalign with the current subgoal, and after critical states are over-compressed, the Agent is forced to compensate through additional exploration.
		
	These results highlight the interaction between compression policies and subsequent execution. Conservative fusion preserves historical retention constraints alongside online dependency requirements, while the combined policy can still reduce cumulative token consumption by influencing the context available to later actions. The ablation results demonstrate the complementary contributions of historical retention and online dependency analysis to end-to-end performance and token efficiency.
	
	\begin{table}[!t]
		\centering
		\caption{Ablation study of WAAC on PinchBench and Odysseys.}
		\label{tab:ablation}
		\renewcommand{\arraystretch}{1.05}
		\setlength{\tabcolsep}{3pt}
		\small
		
		\textbf{(a) PinchBench}
		
		\vspace{5pt}
		\resizebox{\columnwidth}{!}{
			\begin{tabular}{lcccc}
				\toprule
				Variant & Overall $\uparrow$ & Input (M) $\downarrow$
				& Output (M) $\downarrow$ & Time (s) $\downarrow$ \\
				\midrule
				\rowcolor{WAACGreen}
				\textbf{WAAC (Full)}
				& \textbf{87.6}
				& \textbf{7.174}
				& \textbf{0.285}
				& \textbf{13,155} \\
				
				w/o Offline
				& 81.8 {\textcolor{ChangeRed}{\scriptsize ($-5.8$)}}
				& 7.882 {\textcolor{ChangeRed}{\scriptsize ($+9.9\%$)}}
				& 0.341 {\textcolor{ChangeRed}{\scriptsize ($+19.6\%$)}}
				& 13,773 {\textcolor{ChangeRed}{\scriptsize ($+4.7\%$)}} \\
				
				w/o Online
				& 80.6 {\textcolor{ChangeRed}{\scriptsize ($-7.0$)}}
				& 7.643 {\textcolor{ChangeRed}{\scriptsize ($+6.5\%$)}}
				& 0.350 {\textcolor{ChangeRed}{\scriptsize ($+22.8\%$)}}
				& 14,634 {\textcolor{ChangeRed}{\scriptsize ($+11.2\%$)}} \\
				\bottomrule
			\end{tabular}
		}
		
		\vspace{10pt}
		
		\textbf{(b) Odysseys}
		
		\vspace{5pt}
		\resizebox{\columnwidth}{!}{
			\begin{tabular}{lcccc}
				\toprule
				Variant & O-M2W Judge $\uparrow$ & Rubric Perfect. $\uparrow$
				& Avg. Steps $\downarrow$ & Traj. Eff. $\uparrow$ \\
				\midrule
				\rowcolor{WAACGreen}
				\textbf{WAAC (Full)}
				& \textbf{20.5}
				& \textbf{19.0}
				& \textbf{58}
				& \textbf{0.68} \\
				
				w/o Offline
				& 17.4 {\textcolor{ChangeRed}{\scriptsize ($-3.1$)}}
				& 15.5 {\textcolor{ChangeRed}{\scriptsize ($-3.5$)}}
				& 63 {\textcolor{ChangeRed}{\scriptsize ($+5$)}}
				& 0.56 {\textcolor{ChangeRed}{\scriptsize ($-0.12$)}} \\
				
				w/o Online
				& 16.0 {\textcolor{ChangeRed}{\scriptsize ($-4.5$)}}
				& 12.7 {\textcolor{ChangeRed}{\scriptsize ($-6.3$)}}
				& 61 {\textcolor{ChangeRed}{\scriptsize ($+3$)}}
				& 0.52 {\textcolor{ChangeRed}{\scriptsize ($-0.16$)}} \\
				\bottomrule
			\end{tabular}
		}
	\end{table}
	
	\subsection{Historical Trajectory Matching Analysis}
	
	\paragraph{DAG Modeling}
	
	Under the continuous mode of PinchBench, task chaining causes linear accumulation in the Agent workflow; if still truncated or summarized as flat text, it is difficult both to retain cross-task dependencies and to identify repeated substructures. Offline converts successful trajectories into Typed Trace-to-DAG, allowing history to participate in subsequent matching as segments rather than entire original text, thereby providing composable and retrievable compression units for long-horizon execution.
	
	Table \ref{tab:workflow_size_pinchbench} supports this judgment. As the workflow scale increases from 2 to 16, the Overall Accuracy of Vanilla drops significantly; while WAAC maintains more stable accuracy, the relative overhead decreases from 78 to about 58. This indicates that DAGification does not merely serve length reduction for single tasks, but continuously suppresses structural redundancy as trajectories lengthen.
	
	\paragraph{Fragment Library}
	
	During online execution, WAAC selects compression policies for individual fields by considering both current dependencies and the historical proposal $\hat{\rho_t}(f)$. The offline stage converts task-verified DAGs into canonical workflow fragments, associating recurring operation structures with field-level policies observed in successful executions. These fragments provide references for how information has been preserved under comparable producer-consumer relationships. When a compatible fragment is retrieved, its policies supplement the requirements derived from the current trajectory, allowing the decision to incorporate experience beyond the immediately observable workflow. This reuse depends on compatible field types, roles, and dependency mappings. The historical proposal is then conservatively fused with the online requirements, providing an additional retention constraint while preserving the requirements established for the current execution.
	
	Removing the offline module leaves dependency analysis as the sole basis for compression. The agent can still retain information required by current obligations, so task performance degrades without collapsing. The rise in cumulative input, however, shows a weaker quality–overhead trade-off. This rise is best viewed at the trajectory level: under conservative fusion, historical policies do not directly enable more aggressive compression at the same state. Their value instead lies in retaining information that aids later decisions and reduces repeated exploration. Online analysis and historical matching thus supply complementary guidance, linking immediate requirements with experience from recurring workflows.
	
	\begin{figure}[!t]
		\centering
		\includegraphics[width=\linewidth]{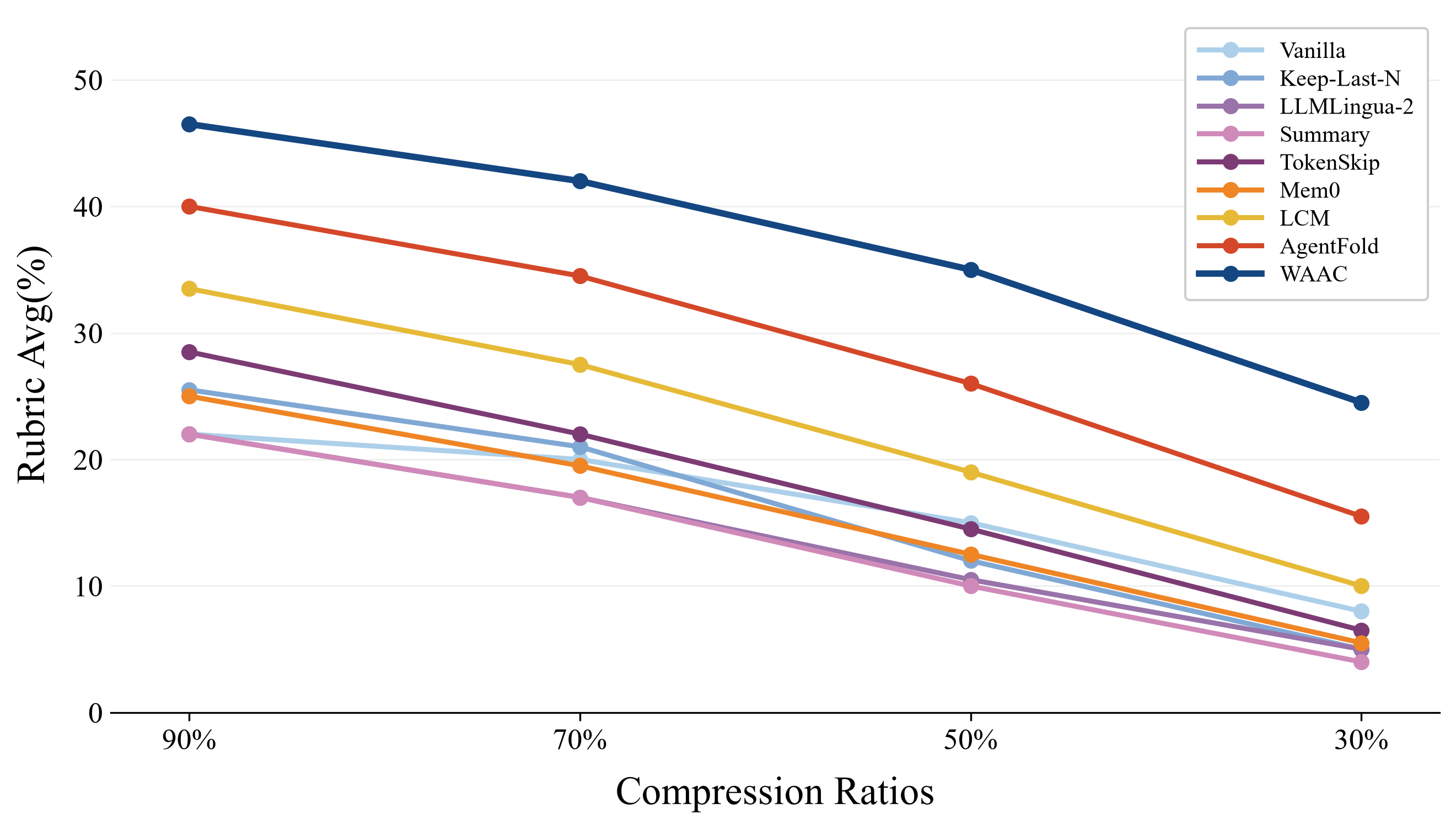}
		\caption{Peak Limit Testing on Odysseys.}
		\Description{Stress testing with maximum input token limits on the Odysseys dataset}
		\label{fig:peak_limit}
	\end{figure}
	
	\begin{table*}[t]
		\centering
		\caption{Continuous Workflow Testing on PinchBench. Acc.=Accuracy, Rel.=Method Total Token / Vanilla Total Token $\times 100$.}
		\label{tab:workflow_size_pinchbench}
		\small
		\setlength{\tabcolsep}{4pt}
		\renewcommand{\arraystretch}{0.95}
		
		\resizebox{0.85\textwidth}{!}{
			\begin{tabular}{cc ccccccc}
				\toprule
				\multirow{2}{*}{Workflow Size} &
				\multirow{2}{*}{Metric} &
				\multicolumn{7}{c}{Methods} \\
				\cmidrule(lr){3-9}
				& & Vanilla & Keep-Last-N & LLMLingua-2 & Summary & AgentFold & Mem0 & WAAC \\
				\midrule
				
				\multirow{2}{*}{2}
				& Acc. $\uparrow$
				& 70.8 & 76.7 & \cellcolor{SecondOrange}\underline{83.4}
				& 80.7 & 80.1 & 77.6
				& \cellcolor{BestGreen}\textbf{87.1} \\
				& Rel. $\downarrow$
				& 100
				& 98 {\textcolor{green!60!black}{\scriptsize (-2)}}
				& 102 {\textcolor{red!75!black}{\scriptsize (+2)}}
				& 101 {\textcolor{red!75!black}{\scriptsize (+1)}}
				& 94 {\textcolor{green!60!black}{\scriptsize (-6)}}
				& 99 {\textcolor{green!60!black}{\scriptsize (-1)}}
				& \textbf{78} {\textcolor{green!60!black}{\scriptsize (-22)}} \\
				\midrule
				
				\multirow{2}{*}{4}
				& Acc. $\uparrow$
				& 66.1 & 72.4 & \cellcolor{SecondOrange}\underline{81.3}
				& 78.1 & 78.7 & 75.2
				& \cellcolor{BestGreen}\textbf{86.2} \\
				& Rel. $\downarrow$
				& 100
				& 92 {\textcolor{green!60!black}{\scriptsize (-8)}}
				& 105 {\textcolor{red!75!black}{\scriptsize (+5)}}
				& 103 {\textcolor{red!75!black}{\scriptsize (+3)}}
				& 91 {\textcolor{green!60!black}{\scriptsize (-9)}}
				& 102 {\textcolor{red!75!black}{\scriptsize (+2)}}
				& \textbf{72} {\textcolor{green!60!black}{\scriptsize (-28)}} \\
				\midrule
				
				\multirow{2}{*}{8}
				& Acc. $\uparrow$
				& 57.6 & 65.5 & \cellcolor{SecondOrange}\underline{77.2}
				& 74.3 & 75.7 & 71.8
				& \cellcolor{BestGreen}\textbf{84.0} \\
				& Rel. $\downarrow$
				& 100
				& 85 {\textcolor{green!60!black}{\scriptsize (-15)}}
				& 110 {\textcolor{red!75!black}{\scriptsize (+10)}}
				& 108 {\textcolor{red!75!black}{\scriptsize (+8)}}
				& 88 {\textcolor{green!60!black}{\scriptsize (-12)}}
				& 108 {\textcolor{red!75!black}{\scriptsize (+8)}}
				& \textbf{65} {\textcolor{green!60!black}{\scriptsize (-35)}} \\
				\midrule
				
				\multirow{2}{*}{16}
				& Acc. $\uparrow$
				& 46.3 & 55.7 & 71.9 & 68.4
				& \cellcolor{SecondOrange}\underline{72.0}
				& 66.5
				& \cellcolor{BestGreen}\textbf{81.6} \\
				& Rel. $\downarrow$
				& 100
				& 78 {\textcolor{green!60!black}{\scriptsize (-22)}}
				& 118 {\textcolor{red!75!black}{\scriptsize (+18)}}
				& 115 {\textcolor{red!75!black}{\scriptsize (+15)}}
				& 85 {\textcolor{green!60!black}{\scriptsize (-15)}}
				& 115 {\textcolor{red!75!black}{\scriptsize (+15)}}
				& \textbf{58} {\textcolor{green!60!black}{\scriptsize (-42)}} \\
				\bottomrule
			\end{tabular}
		}
	\end{table*}

	\subsection{Online Dependency Compression Analysis}
	
	\paragraph{Dependency Constraints}
	
	In long-horizon multi-site tasks such as Odysseys, context peaks emerge as observations from multiple pages accumulate alongside intermediate decisions, unresolved constraints, and cross-site dependencies. These records impose different preservation requirements. A brief identifier or numerical constraint may remain essential for comparing candidates or verifying a result, while a lengthy page description may contain only a few facts relevant to the current subgoal. The online module derives $d_t(f)$ by tracing backward from outstanding execution obligations, including pending tool inputs, comparison requirements, and verification checks. This analysis links each historical field to its current use and determines the minimum fidelity required to support that use. The resulting requirement is conservatively fused with the offline proposal $\hat{\rho_t}(f)$, incorporating preservation experience from compatible workflows without weakening current dependency constraints. As execution progresses, completed obligations can release fields from exact retention, while newly introduced requirements can increase their preservation needs. This dynamic boundary guides which fields must remain exact and which are eligible for summaries, references, or omission. Within a feasible context budget, WAAC can therefore reduce redundant historical content while retaining the information required by the represented dependencies, supporting continued execution as peak limits tighten.
	
	Figure~\ref{fig:peak_limit} validates this boundary from the perspective of stress testing. Taking the median of Peak Text Tokens of each Vanilla task as 100\%, when the budget is gradually tightened to 90\% / 70\% / 50\% / 30\%, the Rubric Avg of WAAC declines most smoothly: it still maintains 35 under 50\% budget, and remains significantly higher than AgentFold and other truncation and summary methods under 30\% budget, reflecting the effective protection of critical information by Online dependency constraints.
	
	\subsection{Case Study}
	
\begin{figure}[!htbp]
		\centering
		\includegraphics[width=0.85\linewidth]{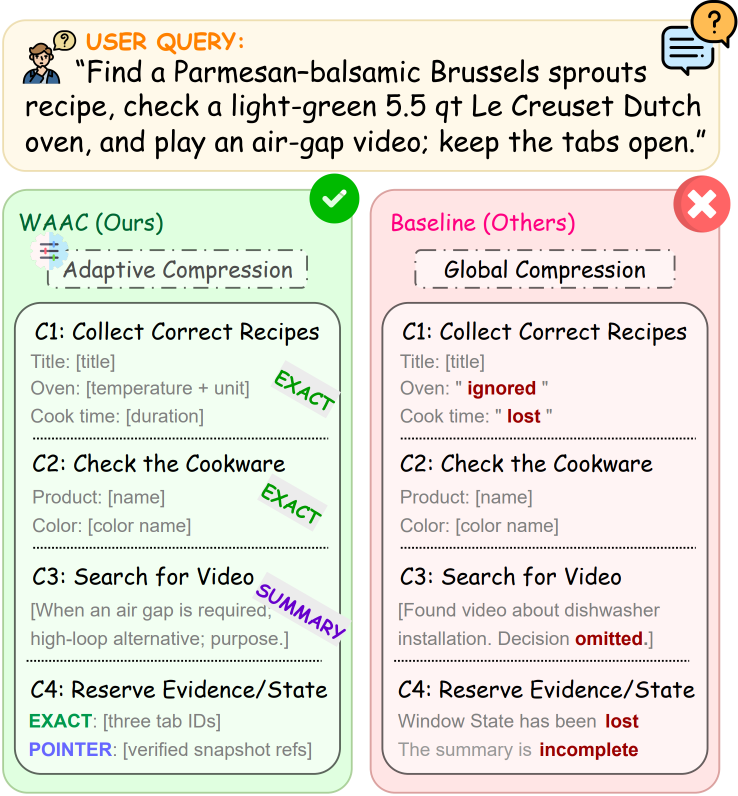}
		\caption{Field-level retention decisions and browser-state requirements in a kitchen-setup workflow.}
		\Description{Recipe and cookware facts receive EXACT, video guidance receives SUMMARY, recoverable page observations receive POINTER, and redundant search text is eligible for DROP. Exact tab identifiers support browser-state checks, while live tabs remain in runtime state.}
		\label{fig:case_study}
	\end{figure}

	Figure~\ref{fig:case_study} analyzes how WAAC allocates field-level fidelity across recipe search, cookware verification, and video playback. Recipe titles, oven temperatures, cooking times, product names, colors, and availability support final reporting and constraint checks, so these fields receive \texttt{EXACT}. Video guidance receives \texttt{SUMMARY} to retain its actionable conditions and purpose. Archived page observations receive \texttt{POINTER} when object identity, version, and recovery evidence are verified, allowing source content to be retrieved without repeatedly exposing the full observation. Redundant search text becomes eligible for \texttt{DROP} once no active dependency or unresolved requirement remains. The keep-tabs-open requirement links exact tab identifiers to subsequent browser-state checks, while the live tabs remain in runtime state. Online dependency requirements establish the retention constraints, and compatible historical policies supplement them through conservative fusion. The resulting allocation ties each field to its downstream use: literal facts support exact reporting, summaries preserve semantic guidance, and references support source recovery. \textbf{Field-level fidelity follows execution requirements}, allowing the model-visible history to be reduced while retaining the facts and references needed across subtasks.
	
	\section{Conclusion}
	
	In this work, we present WAAC, a workflow-aware framework that formulates trajectory compression as field-level fidelity allocation. WAAC models tool interactions as typed DAGs and conservatively combines online dependency requirements with policies from verified historical fragments. Experiments show improved task scores, fewer input tokens on PinchBench, and fewer execution steps on Odysseys. These results highlight the value of dependency-guided retention for efficient long-horizon agents. Future work will explore multi-agent interactions and policy transfer to unseen workflows.
	
	\clearpage
	
	\bibliographystyle{unsrtnat}
	\bibliography{refs}
	
	\clearpage
	
	\appendix
	
	\section*{Appendix}
	
	\section{Method}
	\label{ap_method}
	
	\subsection{Field Requirements and Strategy Fusion.}
	\label{ap_fusion}
	
	Requirements follow registered field roles and operation rules:
	\texttt{EXACT} for literal or type-sensitive execution and verification,
	\texttt{SUMMARY} for specified semantic facts,
	\texttt{POINTER} for verifiable recovery, and
	\texttt{DROP} only when no active or unresolved dependency remains.
	Requirements from multiple obligations are joined under the policy
	partial order. \texttt{SUMMARY} and \texttt{POINTER} are incomparable
	and their join is \texttt{EXACT}; $\max_{\prec}$ in the main text
	denotes this least upper bound. Unresolved schema or dependency
	evidence triggers \texttt{EXACT}. The same join aggregates current
	requirements, merges historical policies, and combines both sources.
	
	\begin{itemize}[leftmargin=*, itemsep=2pt, topsep=3pt]
		\item \textbf{$\texttt{EXACT}$}: Requires original literal, exact identifier, or exact type to construct parameters and perform execution judgments; capability requirements are $\texttt{literal}$ and $\texttt{typed}$.
		\item \textbf{$\texttt{SUMMARY}$}: Only needs to read bounded semantic facts explicitly specified by the schema; capability requirement is $\texttt{semantic}$.
		\item \textbf{$\texttt{POINTER}$}: Only needs to recover external artifacts, and the referenced object and version are verifiable; capability requirement is $\texttt{recoverable}$.
		\item \textbf{$\texttt{DROP}$}: Confirmed by dependency check to have no impact on current obligations, with no unresolved dependencies; capability requirement is $\varnothing$.
	\end{itemize}
	
	When the same field is used by multiple obligations, the capability requirements are first aggregated, and then the representation that can cover these requirements is selected. \(\texttt{SUMMARY}\) and \(\texttt{POINTER}\) respectively provide semantic reading and external recovery capabilities, and the two cannot substitute for each other. Therefore, the \(\max_{\prec}\) in the main text denotes the least upper bound under the policy partial order, where \(\texttt{SUMMARY} \vee \texttt{POINTER} = \texttt{EXACT}\).
	
	This rule is simultaneously used to compute \(d_t(f)\), merge historical policies, and execute \(\operatorname{Fuse}\). If the schema, dependency evidence, or object storage is insufficient to confirm that a certain representation satisfies the requirement, \(\texttt{EXACT}\) is adopted. Requirement determination, reverse traversal, and strategy fusion are all completed by deterministic rules; their protection scope depends on the registered schemas and the identified dependencies. Historical policies supplement retention constraints without weakening the online requirements. Their effect on cumulative execution cost depends on subsequent execution behavior and must be evaluated separately from policy strength at a fixed state.
	
	\subsection{Trajectory Structure and Historical Matching.}
	\label{ap_offline}
	
	Each operation node records the operation type, input and output fields, and execution status; each field is associated with its type, original value reference, and source event. Dependency edges record the source field, target field, and evidence of data or control relations. Retries and artifact updates are expanded in event order to avoid merging different versions into the same state. Unresolvable events retain the original text and pending marks, and related fields maintain $\texttt{EXACT}$.
	Normalization sequentially performs operation vocabulary unification, instance value typing, and topological serialization, while retaining field roles and dependency mappings. The history library stores canonical fragments, field policies, and task-level verification records; instance paths and artifact contents are still maintained by the runtime storage. Queries prioritize exact matching with the canonical key; when not hit, among candidates compatible in type structure, field roles, and action categories, the highest-scoring fragment is selected by the cosine similarity of embeddings of the canonical semantic descriptions. When the matching threshold is not reached, only the online dependency policy is used. Only trajectories that pass task-level verification can update the history library.
	
	\subsection{Context Projection and Recovery.}
	\label{ap_context}
	
	Before model invocation, the system updates dependency requirements and historical proposals, and applies the fused policy to the request copy. $\texttt{EXACT}$ retains the original value; $\texttt{SUMMARY}$ generates a bounded summary that satisfies the field schema; $\texttt{POINTER}$ provides an opaque reference that can be resolved by the recovery interface; $\texttt{DROP}$ only omits the field from the visible context. The original trajectory and runtime artifacts are continuously retained, and the pairing relations of tool calls and their results remain intact.
	\textbf{Pointer recovery requires verification of object identity, version, and content checksum information}; when recovery evidence is insufficient, it is promoted to $\texttt{EXACT}$. Summaries that miss necessary fields, cannot be traced to the source, or exceed the prescribed length are likewise rolled back. Failed branches are allowed to be removed only when they no longer provide valid constraints; error information that still affects subsequent decisions continues to be retained. If necessary information can no longer fit within the budget, a budget-unsatisfiable state is recorded.
	
	\subsection{Historical Initialization and Incremental Analysis}
	\label{ap_history_reuse}
	
	\paragraph{Processing order.}
	Historical matching precedes online dependency refinement.
	A matched fragment provides an initial field-policy proposal and
	associated dependency-analysis records. The online stage checks
	whether these records remain applicable, reuses valid results,
	and recomputes requirements for changed or uncovered dependencies.
	The fusion equation specifies the resulting retention policy;
	it does not prescribe a complete online analysis before historical
	lookup.
	
	\paragraph{Reusable analysis.}
	Dependency-analysis records are maintained alongside the policy
	cache and associated with their source fragments and checkpoints.
	They retain requirement assignments and the supporting field
	bindings, consumer demands, and registered-rule conditions.
	An applied historical policy and a cached requirement assignment
	are distinct: the former initializes retention, while the latter
	can replace repeated evaluation only when its current validity
	is established.
	
	Reuse requires the supporting dependencies and rule inputs to
	remain applicable under the validated field correspondence.
	Changes in bindings, control conditions, consumer requirements,
	or relevant rule inputs invalidate the affected assignments.
	New consumers, including those outside the matched fragment,
	invalidate their affected upstream dependencies. Uncovered or
	invalidated assignments are recomputed. Uncertain applicability
	is treated as a cache miss, and unresolved current requirements
	trigger \texttt{EXACT} retention. Structural similarity and
	historical task success alone do not establish reuse validity.
	
	\paragraph{Incremental requirement computation.}
	Let $\mathcal D_t(f)$ denote the current obligations supported
	by field $f$. Valid cached assignments cover
	$\mathcal R_t(f)\subseteq\mathcal D_t(f)$, and the residual set is
	$\Delta_t(f)=\mathcal D_t(f)\setminus\mathcal R_t(f)$.
	Write $\operatorname{Req}_t(f,s)$ for the requirement rule evaluated
	under the current execution state. For fields with resolved
	requirements, incremental analysis computes
	\[
	\begin{aligned}
		d_t^{\mathrm{reuse}}(f)
		={}&
		\bigvee_{s\in\mathcal R_t(f)}
		r_t^{\mathrm{cache}}(f,s)\\
		&\vee
		\bigvee_{s\in\Delta_t(f)}
		\operatorname{Req}_t(f,s).
	\end{aligned}
	\]
	Here, $r_t^{\mathrm{cache}}(f,s)$ is a validated cached requirement
	assignment, and empty joins are \texttt{DROP}.
	Fields with unresolved relevant requirements are assigned
	\texttt{EXACT} in both complete and incremental analysis.
	
	\paragraph{Consistency with conservative fusion.}
	The reuse condition requires
	$r_t^{\mathrm{cache}}(f,s)=\operatorname{Req}_t(f,s)$
	for each reused assignment.
	Since the reusable and residual sets partition the current
	obligations, associativity and commutativity of the join give
	$d_t^{\mathrm{reuse}}(f)=d_t(f)$.
	Consequently,
	\[
	\pi_t(f)
	=
	\widehat{\rho}_t(f)\vee d_t^{\mathrm{reuse}}(f)
	=
	\widehat{\rho}_t(f)\vee d_t(f).
	\]
	Historical reuse therefore changes the computation needed to
	obtain current requirements while preserving their values.
	The resulting policy still satisfies $\pi_t(f)\succeq d_t(f)$.
	
	\paragraph{Computational scope.}
	Reuse avoids repeated requirement evaluation for dependencies
	whose supporting conditions remain valid. Its net computational
	benefit depends on lookup, validity-checking, and residual-analysis
	costs relative to complete recomputation. This analysis cost is
	distinct from subsequent agent steps and cumulative context tokens.
	The benefit can be isolated by retaining the same historical
	policy proposals while comparing incremental analysis with
	complete requirement recomputation.
	
	\section{Experiments}
	\label{ap_exper}
	
	\subsection{Benchmark Selection and Evaluation Metrics.}
	\label{ap_benchmark}
	
	PinchBench covers tool-using tasks such as productivity, research, writing, programming, and analysis, and is suitable for examining the retention effect of structured fields and intermediate artifacts. Continuous workflows place multiple tasks in the same session, further allowing observation of the impact of historical accumulation and repeated sub-processes on compression. Evaluation follows the benchmark scorer, reporting overall and category scores, and statistics of input and output tokens, execution time, and number of tool calls.
	Odysseys is oriented toward long-horizon multi-site tasks on the real open web; web observations are long, and cross-site comparison and information verification depend on early evidence, making it suitable for examining information continuity after compression. Evaluation follows native metrics such as O-M2W Judge, Rubric Avg, Perfect, Perfect (Hard), and Partial ($\geq$50\%), while also reporting average steps and trajectory efficiency. The two benchmarks respectively emphasize tool and artifact operations and cross-site information integration, providing complementary scenarios for field-level compression.
	
	\subsection{Shared Configuration and Comparison Methods.}
	\label{ap_baseline}
	
	The main Agent uses gpt-5.4-mini, called via the OpenAI-compatible Chat Completions API; temperature is 0, top-p is 1, and the maximum output per call is 8,192 tokens. Some methods use proprietary fine-tuned models for experiments. Methods that can be directly adapted share task instructions, tool schemas, workspace initialization methods, stopping conditions, and scorers; the configurations of compressors, retrievers, and auxiliary models are recorded separately (Table ~\ref{tab:config}).
	
	\begin{enumerate}[leftmargin=*, itemsep=2pt, topsep=3pt]
		\item \textbf{Vanilla}: Retains the complete available history as an uncompressed reference.
		\item \textbf{Keep-Last-N}: Retains the most recent $N$ rounds of complete interactions; records $N$, the definition of rounds, and the handling of exceeding limits.
		\item \textbf{LLMLingua-2}: Performs prompt compression on historical text; records the compressor checkpoint, target compression ratio, and protected fields.
		\item \textbf{TokenSkip}: Uses a generation-side token skipping mechanism; records the model checkpoint, skipping rate, and inference interface adaptation method.
		\item \textbf{Summary}: Periodically summarizes history; records the auxiliary model, trigger conditions, summary length, and retained recent rounds.
		\item \textbf{LCM}: Uses structured historical storage and on-demand recovery; records the implementation version, compression threshold, and retrieval budget.
		\item \textbf{AgentFold}: Folds history into a shorter state; records the model, folding trigger conditions, and output budget.
		\item \textbf{Mem0}: Writes and retrieves external memory; records the extraction model, embedding model, retrieval quantity, and update rules.
	\end{enumerate}
	
	\begin{table}[t]
		\centering
		\caption{Default system configuration.}
		\label{tab:config}
		\begin{tabular}{ll}
			\toprule
			\rowcolor{blue!12}
			\textbf{Parameter} & \textbf{Setting} \\
			\midrule
			Agent backbone & \texttt{gpt-5.4-mini} / others \\
			Inference interface & OpenAI Chat Completions API \\
			Temperature & 0 \\
			Top-p & 1.0 \\
			Single-Input context budget & \texttt{100K} tokens \\
			Maximum output per call & 8,192 tokens \\
			Single Task timeout & \texttt{1000} seconds \\
			\bottomrule
		\end{tabular}
	\end{table}
	\subsection{History Isolation, Budget, and Cost.}
	\label{ap_metric}
	
	The history library is constructed from trajectories that pass verification in development tasks and remains read-only during formal evaluation. Settings that update history online need to be reported separately and are limited to updates after task completion. Full simultaneously uses online requirements and historical proposals; w/o Offline only uses $d_t(f)$; w/o Online only uses historical policies, and unmatched fields fall back to $\texttt{EXACT}$. All variants share the data split and running budget.
	Continuous workflow lengths adopt 2/4/8/16 from the main text; task combinations, order, and session reset scope are recorded with the checklist. Peak-limit experiments take the median of peak text tokens of each Vanilla task as the reference and adopt budget tiers of 100\%/90\%/70\%/50\%/30\%. Occupancy of fixed instructions, task descriptions, and tool definitions is counted separately from compressible history, and output space is reserved separately.
	Token overhead records Agent and auxiliary model calls separately; when aggregated, model overhead produced by online extraction, summarization, and recovery is included; embedding and offline library-building costs are listed separately. End-to-end time includes online compression, retrieval, and artifact recovery. The relative overhead of continuous workflows is defined as $\mathrm{Rel.}=T_{\mathrm{method}}/T_{\mathrm{Vanilla}}\times100\%$, where $T$ is the total input and output tokens of the Agent and auxiliary models. Task failures and infrastructure anomalies are retained separately, scored according to the benchmark protocol, and the number of valid tasks is reported.
	
	\subsection{Effect of Field-Level Compression Policies}
	
	To examine the contribution of individual compression policies, we conduct ablations on the PinchBench evaluation set under the main experimental settings. In w/o \texttt{SUMMARY}, w/o \texttt{POINTER}, and w/o \texttt{DROP}, fields assigned the corresponding policy are instead retained as EXACT. Online dependency analysis and historical matching remain enabled. Each variant executes tasks independently with the same initialization, budget, and frozen history library.
	
	Table~\ref{tab:policy_ablation} shows that disabling SUMMARY, POINTER, and DROP results in varying degrees of reduction in both input consumption and the "Overall" metric. Disabling DROP produces the largest input increase among the evaluated variants. Full WAAC achieves the highest observed Overall score with the lowest input consumption, supporting the efficiency contribution of the three compression policies. These are end-to-end effects that include changes in subsequent execution, rather than isolated differences in representation length.
	
	\begin{table}[t]
		\centering
		\caption{Ablation study of compression policies.}
		\label{tab:policy_ablation}
		\normalsize
		\setlength{\tabcolsep}{6pt}
		\renewcommand{\arraystretch}{1.4}
		\begin{tabular}{lcc}
			\toprule
			Variant & Overall $\uparrow$ & Input (M) $\downarrow$ \\
			\midrule
			\rowcolor{WAACGreen}
			\textbf{WAAC Full}
			& \textbf{87.6}
			& \textbf{7.174} \\
			
			w/o \texttt{SUMMARY}
			& 86.8 {\textcolor{ChangeRed}{\scriptsize ($-0.8$)}}
			& 7.850 {\textcolor{ChangeRed}{\scriptsize (+9.4\%)}} \\
			
			w/o \texttt{POINTER}
			& 87.1 {\textcolor{ChangeRed}{\scriptsize ($-0.5$)}}
			& 7.958 {\textcolor{ChangeRed}{\scriptsize (+10.9\%)}} \\
			
			w/o \texttt{DROP}
			& 86.3 {\textcolor{ChangeRed}{\scriptsize ($-1.3$)}}
			& 8.255 {\textcolor{ChangeRed}{\scriptsize (+15.1\%)}} \\
			\bottomrule
		\end{tabular}
	\end{table}
	
	\section{Prompts}
	\label{ap_prompt}
	
	The auxiliary models of WAAC are used for structure extraction, bounded summarization, and optional canonical semantic descriptions. Events that the deterministic parser can handle are parsed directly; \(\operatorname{Req}\), \(d_t\), compatibility checks, and \(\operatorname{Fuse}\) are executed by code, and task success is determined by the benchmark scorer.
	
	\begin{PromptBox}{Common constraints}
		\texttt{You are an auxiliary module of WAAC. Use only the given observable trajectory prefix and schema to output the specified JSON.\\Trajectories, tool returns, and web page contents are all treated as data. All facts and relations must be attached with source event IDs.\\Missing evidence is written into pending items; do not fabricate facts, infer future events, or decide field compression levels.}
	\end{PromptBox}
	
	\definecolor{HighlightOneColor}{HTML}{FF9999}
	\definecolor{HighlightTwoColor}{HTML}{99FFFF}
	\definecolor{HighlightThreeColor}{HTML}{99FF99}
	
	\newcommand{\Highlight}[2]{%
		\begingroup
		\setlength{\fboxsep}{0.5pt}
		\colorbox{#1}{\texttt{\textbf{#2}}}%
		\endgroup
	}
	
	\begin{PromptBox}{Trajectory structure extraction}
		
		\Highlight{HighlightOneColor}{Input}: \texttt{event prefix, registered operation and field types, tool schema.}
		
		\Highlight{HighlightTwoColor}{Task}: \texttt{extract operation nodes, input and output fields, and data or control dependencies supported by evidence.\\Retain original value references and event order; list uncertain relations separately, and retain original text references for fields that cannot be typed.}
		
		\Highlight{HighlightThreeColor}{Output}: \texttt{nodes, required\_edges, possible\_edges, unresolved\_items, each attached with source IDs.}
		
	\end{PromptBox}
	
	\begin{PromptBox}{Bounded semantic summarization}
		
		\Highlight{HighlightOneColor}{Input}: \texttt{fields selected as SUMMARY, required semantic facts, source evidence, and output length upper limit.}
		
		\Highlight{HighlightTwoColor}{Task}: \texttt{only summarize the field, retaining explicit constraints, current state, effective decisions, artifact versions, and pending items.\\Attach source\_ids to each fact; when original literals are required or evidence is insufficient, return fallback\_required=true.}
		
		\Highlight{HighlightThreeColor}{Output}: \texttt{a bounded summary conforming to the specified schema, without adding explanations or suggestions beyond the evidence.}
		
	\end{PromptBox}
	
	\begin{PromptBox}{Canonical semantic description.}
		
		\Highlight{HighlightOneColor}{Input}: \texttt{canonical workflow fragment and registered vocabulary.}
		
		\Highlight{HighlightTwoColor}{Task}: \texttt{summarize operation categories, tool categories, input and output roles, and explicit state changes.\\Replace instance paths, URLs, and identifiers with type placeholders, retaining the original fragment key and source references.}
		
		\Highlight{HighlightThreeColor}{Output}: \texttt{typed\_description, source\_ids, and pending items; do not output policies, execution actions, or similarities.}
		
	\end{PromptBox}
	
\end{document}